\documentclass[conference]{IEEEtran}
\IEEEoverridecommandlockouts
\usepackage{cite}
\usepackage{amsmath,amssymb,amsfonts}
\usepackage{algorithmic}
\usepackage{graphicx}
\usepackage{textcomp}
\usepackage{xcolor}
\usepackage{todonotes,multirow,enumitem,hyperref,flushend}
\usepackage{latexsym}
\usepackage{longtable}

\def\BibTeX{{\rm B\kern-.05em{\sc i\kern-.025em b}\kern-.08em
    T\kern-.1667em\lower.7ex\hbox{E}\kern-.125emX}}

\begin{document}

\title{\textit{What Do Patients Say About Their Disease Symptoms?} Deep Multilabel Text Classification With Human-in-the-Loop Curation for Automatic Labeling of Patient Self Reports of Problems
}

\author{\IEEEauthorblockN{Lakshmi Arbatti, Abhishek Hosamath, Vikram Ramanarayanan and Ira Shoulson}
\textit{Modality.AI, Inc.}\\
\texttt{<lakshmi.arbatti, abhishek.hosamath, v, ira.shoulson>@modality.ai}}
   

\maketitle

\begin{abstract}
The USA Food and Drug Administration has accorded increasing importance to patient-reported problems in clinical and research settings. In this paper, we explore one of the largest online datasets comprising 170,141 open-ended self-reported responses (called `verbatims') from patients with Parkinson’s (PwP) to questions about what bothers them about their Parkinson’s Disease and how it affects their daily functioning, also known as the Parkinson’s Disease Patient Report of Problems\textsuperscript{\textcopyright}. Classifying such verbatims into multiple clinically relevant symptom categories is an important problem and requires multiple steps -- expert curation, a multi-label text classification (MLTC) approach and large amounts of labelled training data. Further, human annotation of such large datasets is tedious and expensive. We present a novel solution to this problem where we build a baseline dataset using 2,341 (of the 170,141) verbatims annotated by nine curators including clinical experts and PwPs. We develop a rules based linguistic-dictionary using NLP techniques and graph database-based expert phrase-query system to scale the annotation to the remaining cohort generating the machine annotated dataset, and finally build a Keras-Tensorflow based MLTC model for both datasets. The machine annotated model significantly outperforms the baseline model with a F1-score of 95\% across 65 symptom categories on a held-out test set.
\end{abstract}


\section{Introduction}

Healthcare and clinical research is an information intensive industry \cite{wilcox2003role}. The advent of Electronic Health Records (EHRs) and the availability of large amounts of clinical notes has piqued the interest of many researchers and advanced the field of Natural Language Processing (NLP). In order to automate and properly analyze and process available information, data needs to be extracted from such large corpora and arranged in a structured form understandable to computers. Classification of such information into structured reports and labels is one of the most common approaches to medical text analytics.  Several pre-trained language models have been built, trained on these clinical notes and on several million other data points. These NLP algorithms have historically been used to perform such classification. One such example is the classification of free-text triage of chief complaints in pre-determined syndromic categories \cite{chapman2005classifying}. However, \textit{clinically-relevant} classification requires expert knowledge in order to extract domain specific information and descriptors from within free text. There is no one size fits all or off the shelf solution to text analytics. An ensemble of approaches is therefore necessary to explore and classify data specific to the use case. A human-in-the-loop approach ensures the results of the system are validated for accuracy and clinical relevance throughout the process of model training including generation of a gold standard test set, data scaling and model validation.


\subsection{Background and PD-PROP\textsuperscript{\textcopyright} Data Collection}

The voice of the patient has been accorded increasing research and regulatory attention, largely catalyzed by disease-focused advocacy organizations, enactment of the Twenty-First Century Cures Act\footnote{Office of the Commissioner. 21st Century Cures Act. FDA: \url{https://www.fda.gov/regulatory-information/selected-amendments-fdc-act/21st-century-cures-act}}, and the FDA Patient-Focused Drug Development (PFDD) initiative\footnote {CDER Patient-Focused Drug Development. FDA: \url{https://https://www.fda.gov/drugs/development-approval-process-drugs/cder-patient-focused-drug-development}}. What patients report about their illness is of critical importance, but has traditionally been captured using categorical scales that are rated by clinicians in research settings. Obtaining patient verbatim reports directly has not been considered feasible because of wide inter-patient variability and lack of quantification methods. However, when a patient is asked especially in a confidential online setting about what bothers them most about their disease, the responses elicited are far more nuanced and insightful compared to a face-to-face interaction with their clinical specialist that typically averages about 27 seconds and may be biased towards clinician expectations. The advent of online research platforms and maturation of medical informatics have enhanced the systematic capture and analysis of what patients experience or feel. FoxInsight (FI)\footnote{Fox Insight: \url{https://foxinsight.michaeljfox.org}} is one of the largest ongoing online study platforms funded by Michael J Fox Foundation for Parkinson's Research (MJFF) \cite{smolensky_fox_2020} that collects from consenting adults diagnosed with PD, among other questionnaires, the PD-PROP \cite{javidnia_predictive_2021,shoulson_longitudinal_2022}. 
Similar studies applying PROP\textsuperscript{\textcopyright} are currently underway for Huntington's Disease, Multiple Sclerosis, Depression, etc. Hence, it is imperative that process methodology (including a pre-trained model) be created that can help classify such problem reports when applied in different disease and research settings.

\subsection{Related Work}
There are several use cases of MLTC such as genre detection \cite{9668427}, topic modelling \cite{nawab_natural_2020} \cite{8489513}, plain medical text mining within electronic health records (EHR) \cite{zhang2018multi}. Deep learning vector embedding algorithms such as Doc2Vec \cite{8489513}, Universal Sentence Encoder \cite {universal_sentence_encoder} and so on are powerful tools that can detect document similarities in a large vector space. However, they stop short in that the resulting document similarities need to be manually evaluated in order to glean additional insights from category clusters. FasTag approach to automated annotation of clinical records to match ICD-9 and ICD-10 codes for billing has yielded reasonable accuracy for veterinary data (91\%) however results have been lower (71\%) for human records \cite{venkataraman_fastag_2020}. Using pre-trained models such as BERT \cite{bert_2018} result in label classifications that are more generic since they cater to multiple input data types such as EHR data or clinical notes \cite{turner_information_2022}. Moreover, categorizing verbatims into specific clinical symptom categories is challenging as they can be very nuanced such that different people could effectively report the same symptom in different ways. Besides, using generic pre-trained models require significant data resources and computational capabilities in the training phase \cite{pranjic2020evaluation}. In the past, traditional rules-based techniques have yielded the best performance when it comes to domain heavy classification problem. A rule-based dictionary structure is simple to use and easy to implement, however they do not perform well when unknown entities are encountered and tend to result in low recall since the rules cater to very specific data sets \cite{9568778}. 
To the best of our knowledge, there are no tools that both capture and automatically label patient report of problems according to different categories of symptoms in a clinically meaningful manner. 
\section{Methodology}
\begin{figure*}[htb]
  \centering
  \includegraphics[width=\textwidth]{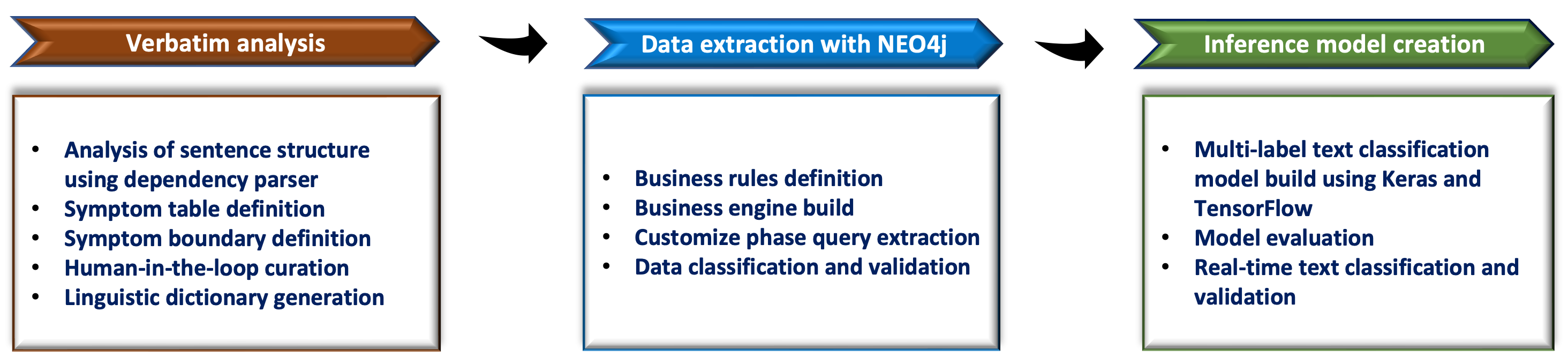}
  \caption{Flowchart depicting the various phases in the process}
  \label{fig:flowchart}
\end{figure*}

Our methodology comprises of a three-step approach as described in Figure \ref{fig:flowchart}. The first step includes an initial analysis of the verbatims to provide the curation team with the knowledge they need to define disease specific symptoms and domains. This also helps define a rules-based process to build a linguistic dictionary comprising of synonyms and similar terms and phrases. In the second step, we apply the rules and generated linguistic dictionary to scale the data across the entire cohort and finally in the third step, we train a deep learning model to perform MLTC. 

\subsection{Verbatim Analysis}
\begin{figure*}[htb]
\centering
    \includegraphics[width=1\textwidth,height=0.4\textheight]{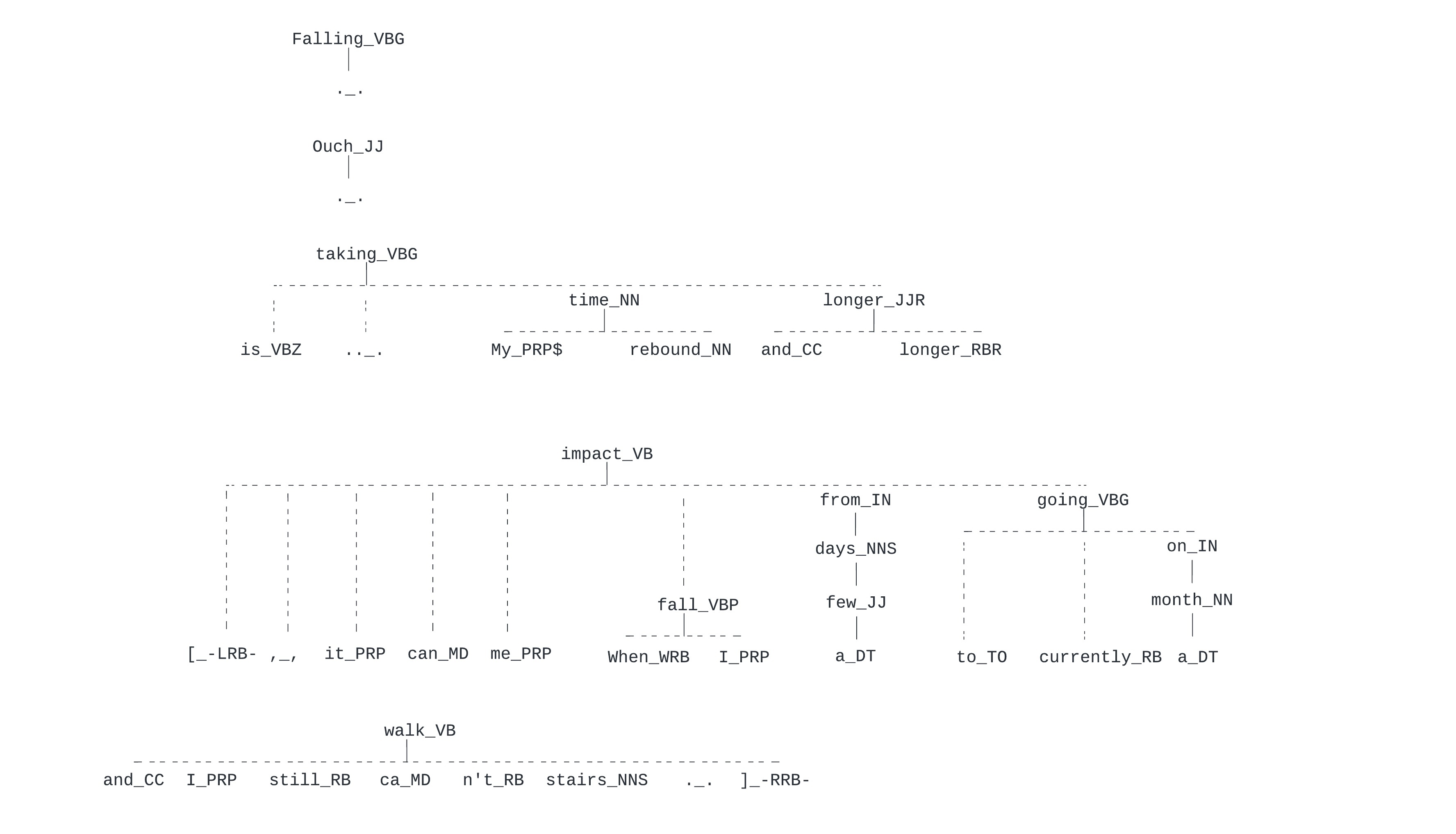}
    \caption{Dependency tree depicting the parts of speech of the sample verbatim}
    \label{fig:dependency_tree_parser}
\end{figure*}

For simplicity, the PROP question “What is the most bothersome problem for you due to your Parkinson's disease?” is referred to as the \textbf{\textit{problem}} and “In what way does this problem bother you (by affecting your everyday functioning or ability to accomplish what needs to be done)?” as the \textbf{\textit{consequence}}. We analyze a concatenated report of the problem and consequence that we henceforth term a “verbatim”. We observed that analyzing either the problem or the consequence as a stand alone was not as informative compared to the whole verbatim. 
As an initial approach to understanding the linguistic sentence structure of verbatims, we examined several of them through dependency parsing to evaluate a pattern of commonality in the sentence structure (Figure \ref{fig:dependency_tree_parser}). The responses comprised predominantly of a combination of verbs, nouns and adjectives. 

\subsubsection {Human-in-the-loop Curation}
Next, the curators discussed amongst themselves typical problems and symptoms experienced by PwP, and came up with a list of 14 domains (broad area of PD symptoms), with each domain comprising of several specific symptoms totalling 65. As an example of domain and symptom definition, Postural Instability was defined as a domain while balance, falling and fear of falling were defined as symptoms within the Postural Instability domain. Of the 14 domains defined, 8 were motor related and 6 were non-motor related, including cognitive, psychiatric, sleep, autonomic dysfunction, fatigue, and pain. This distinction is important since the model developed can independently classify non-motor symptoms which are common and ubiquitous across different diseases. For a more detailed list of symptoms, refer to Table \ref{tab:validation_metrics}.

\noindent The outcome of this exercise resulted in a symptom definition table that not only defined the symptoms to be classified, but also delineated the inclusion and exclusion criteria for each of the 65 symptoms. The symptom definition table provides a reference for a robust clinically validated understanding of the defined symptoms, thereby bridging the gap among researchers and the user community at large in interpreting the predicted outcome labels of the deep learning model, a sample of which is depicted in (Table \ref{tab:symptomtable}). 

\begin{table*}[h]
\centering
\caption{Symptom definition table for RBD like symptoms with sample phrases.}
\begin{tabular}{ |p{1cm}|p{2cm}|p{5cm}|p{3cm}|p{3cm}|  }
 \hline
 \multirow{2}{*}{Domain} &   \multirow{2}{*}{Symptom} & \multicolumn{2}{|c|}{Conceptual Boundaries} & \multirow{2}{*}{Sample Phrases} \\
 \cline{3-4}
    &   &   Includes    &   Excludes    &   \\
 \hline
Sleep & REM sleep behavior disorder (RBD) like symptoms & Acting out dreams, movements, or vocalizations during sleep; may only be aware due to partner report; explicit reports of REM sleep issues; night terrors &
Moving in sleep without further specification & acting out dreams, RBD, REM sleep behavior disorder, yelling in sleep \\
 \hline
\end{tabular}
\label{tab:symptomtable}
\end{table*}

\noindent About 2,341 (up to 50 verbatim samples per symptom category) of the 170,141 verbatims, i.e., 1.4\% were curated by experts, who labelled the symptoms and provided terms or phrases within the verbatim that were indicative of this classification.

\subsubsection{Generation of Linguistic Dictionary}

Our next step in the process was to build a linguistic dictionary\footnote{\url{https://plato.stanford.edu/archives/win2022/entries/natural-language-ontology/}} using the knowledge of the sentence structure and the curation output. The dictionary was constructed using a three-step approach:
\begin{enumerate}[noitemsep,nolistsep]
    \item Parts of speech extraction - We used parts of speech, specifically nouns, adjectives and verbs to provide an exploratory visualization of the various aspects of symptom reporting to aid the curators to define the domains and symptoms.
    \item \textit{word2vec} model trained on clinical trials and pubmed data for synonym detection - This phase was an important part of building the linguistic dictionary for each of the specific terms. For example, when our word2vec model was queried to provide 4 terms that had the highest probability of being similar to "dystonia" a condition mentioned by patients, the model was able to correctly identify certain synonyms such as cramping, calf, ankle as other terms commonly used in a context similar to those reporting dystonia as their bothersome problem (Table \ref{tab:word2vec}).
    \item Unified Medical Language System (UMLS)\footnote{\url{https://www.nlm.nih.gov/research/umls/index.html}} controlled unique identifier (CUI) extraction to develop a comprehensive table of all possible terms \cite{sabbir_knowledge-based_2017} and phrases that patients would typically use to describe a specific symptom thereby resulting in the linguistic dictionary that could then be passed into an expert system to extract appropriate verbatims to scale. See Table \ref{tab:cuirelationshiptable} for examples of the UMLS CUI extraction. Take for example the verbatim: \textit{"depression, very unhappy about movement problems, fear of the future. [I can't do the things I used to easily do]"}. In the above example, we see that related terms such as "Unhappiness (C0476477)" were used to build the linguistic dictionary for Depressive Symptoms (C0086132). 
\end{enumerate}


\begin{table*}[h]
\centering
\caption{Synonyms and similarity probabilities obtained when the word2vec model was queried to provide 4 terms that had the highest probability of being similar to "dystonia" a condition commonly mentioned by patients.}
\begin{tabular}{ |p{2cm}|p{1.5cm}|p{11cm}|  }
 \hline
\textit{word2vec} synonym & Similarity prob. & Exemplar verbatim classified by proposed algorithm \\
 \hline
'calf'  &   0.776   &   Left side calf contraction with toe curling. [Difficult to drive and walk occasionally.] \\
'neuropathy'    &   0.776   &   Loss of muscle mass due to poor muscle nerve stimulation, curling of toes and neuropathy in feet from tight muscles in legs and buttocks. [It is mainly an annoyance. Toe curl leads to hammertoes, problem getting comfortable supportive shoes.]\\
'ankle' &   0.743   &   Pain, ankle pain resulting from dystonia, shoulder pain and intermittent pain from leg cramps. [Ankle pain makes walking difficult. Shoulder pains restricts upper body movement. Leg cramps interrupt my already fractured sleep. ]\\
'dyskinesia'    &   0.729   &   Dyskinesia. [Leads to dystonia in neck and shoulder, cramping toes] \\
 \hline
\end{tabular}
\label{tab:word2vec}
\end{table*}

\begin{table*}[h]
\centering
\caption{Example of CUI relationships}
\begin{tabular}{|p{2cm}|p{2cm}|p{2cm}|p{2cm}|p{2cm}|p{2cm}|  }
 \hline
 Symptom CUI & \multicolumn{5}{|c|}{Related CUIs} \\
 \hline
Depressive Symptoms (C0086132) & Anhedonia (C0178417) & Dysphoric mood (C0233477) & Melancholia (C0025193) & Unhappiness (C0476477) & Depressed mood (C0344315) \\
 \hline
\end{tabular}
\label{tab:cuirelationshiptable}
\end{table*}

\begin{table*}[h]
\caption{Term table for RBD like symptoms.}
\centering
\begin{tabular}{ |p{1cm}|p{1.5cm}|p{0.8cm}|p{1.5cm}|p{1.5cm}|p{1.5cm}|p{1.5cm}|p{1.5cm}|p{1.5cm}|  }
 \hline
 Domain &   Symptom &   serial  &   term1   & term2   & term3   & term4   & term5   & term6   \\
 \hline
Sleep &  RBD like symptoms  &   1   &   enact my dream  &   active dream    &   fighting in my sleep   &   act out near the term nightmare    &   scream AND sleep    &   thrash near the term dream\\

 \hline
\end{tabular}
\label{tab:termtable}
\end{table*}

\subsection{Data extraction with NEO4j}

\begin{figure}[htb]
\centering
    \includegraphics[width=80mm]{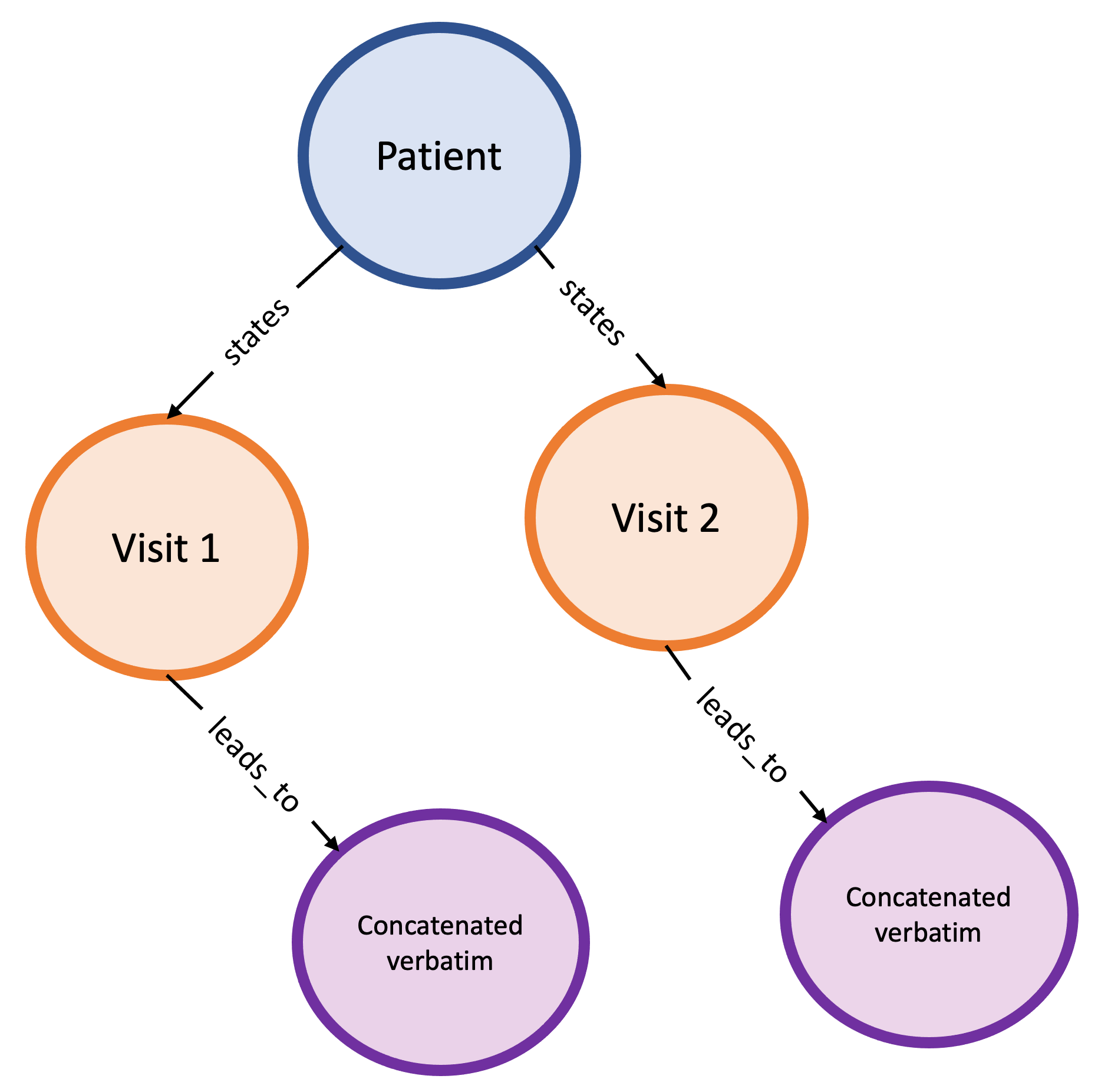}
    \caption{Graph representation of the verbatim database}
    \label{fig:graph_representation_neo4j}
\end{figure}

NEO4j’s\footnote{\url{https://neo4j.com/}} full-text search engine powered by Apache Lucene was used along with custom search rules \cite{fries_swellshark_2017,wang_clinical_2019,cusick_using_2021} to scale the annotation from the curated 2,341 verbatims to the entire dataset of over 170,141 verbatims \cite{venkataraman_fastag_2020,zhang_recent_2020,mohammad_extraction_2022}. The entire dataset comprising verbatims and all other participant and visit-related information was organized in Neo4j’s graph schema that uses nodes and relationships to store data. Indexes were set on unique identifying information such as ParticipantID for high database operational performance\footnote{\url{https://neo4j.com/docs/cypher-manual/current/indexes-for-search-performance/}}. Study-level information such as demographics was stored in \textit{patient nodes}, under which multiple linked \textit{visit nodes} were created to store visit-level information such as visit number and verbatim reports for each visit. \textit{Extension nodes} for each visit node were further created and indexed to store the concatenated verbatim problem and consequence reports for high-speed querying capabilities. All nodes were linked to each other through relationships (Figure \ref{fig:graph_representation_neo4j}). Methodical rules were developed for the annotation of symptoms labels using full-text search capabilities such as single-term query, phrase query, wildcard query, range query, regex query, fuzzy query, etc\footnote{\url{https://lucene.apache.org/core/}}. These rules were created for each symptom and stored in term tables based on the following:
\begin{itemize}[noitemsep, nolistsep]
\item Symptom Inclusion and Exclusion criteria provided by curators (PD clinicians and experienced experts)
\item Annotation and terms/phrases provided by curators
\item Closely related terms derived from the algorithm defined in the previous section, which was developed using word2vec classifications, UMLS and CUIs
\item ICD-10 codes. 
\end{itemize}
\noindent Table \ref{tab:termtable} lists an example term table for REM sleep behavior disorder (RBD).

\noindent A Python script was used to connect to the Neo4j database for annotation and extraction of the verbatim data. Queries were used in conjunction with the rules to loop through all verbatims and annotate them. The resulting data was further reshaped with unclassified and blank rows removed using Pandas\footnote{\url{https://pandas.pydata.org}} to obtain a comprehensible analytical dataset comprising verbatims and their associated symptoms. Approximately 1\% of the analytical dataset was randomly sampled for each symptom, for validation by the curators. To further challenge the capability of the rules and machine classification, symptoms in the validation dataset were enhanced with “negatives” which were non-symptoms but closely related to the symptom being validated (e.g., the sleep symptom samples were enriched with verbatims predicted to report fatigue). The following are example verbatims of closely related symptoms: 
\begin{itemize}[noitemsep,nolistsep]
    \item \textit{Internal tremor: "slowness in doing things, internal quivering. [it's not something I had to deal with before PD]"}
    \item \textit{Anxiety: "Anxiety - generalized. [Makes it hard to speak publicly and voice quivers]"}
\end{itemize}

\noindent The results of the validation were then compared to the machine classification, after which the constructed rules were further optimized/fine-tuned for symptoms with lower accuracy. The optimized model was able to annotate verbatims with an accuracy of 96-100\% for various symptoms from 1555 verbatims as manually validated by curators (Table \ref{tab:validation_metrics}). 
This dataset was further used for training a scalable machine-learning model.

\subsection{Inference Model creation / Keras Multi-Label Text Classification Model}

Of the total 170,141 available verbatim samples, 2,341 were annotated by humans. About 445 of uniformly distributed samples from this set was then set aside as the held-out test set for model evaluation. The remaining 1896 samples, constituted the "baseline" training set. The "baseline model" was trained using this set. The second evaluation set was created from the remaining 167,800 samples that were annotated by our described process. The trained model on this set constituted the "machine annotated model". Both models were trained using the same Tensorflow Keras\footnote{\url{https://keras.io}} deep learning model architecture.

\noindent When building the machine annotated model, of the unique label classifications of 10,519 unique multi label combinations, rare label classifications i.e., label combinations with frequency of 1 were eliminated from the dataset to remove class imbalance, resulting 159,115 verbatims available for model training.

\noindent A train-test split of 90-10 was used. The test set was further split into test and validation sets at 50-50 ratio. Sklearn’s\footnote{\url{https://scikit-learn.org}} train-test-split was used to achieve this task resulting in 143,203 reports in the training set, 7956 test and 7956 validation samples. The data was pre-processed by multi-hot encoding using Keras' \textit{StringLookup} function and transformed into vectors using the Keras \textit{TextVectorization} function, which transformed data into bi-grams first and then represented them using TD-IDF (term frequency-inverse document frequency).

\noindent The deep learning model architecture consisted of two fully connected hidden layers with 512 and 256 neurons and one output layer was then constructed. Among the several combinations of activation functions that were run, a combination of ReLU (Rectified Linear Unit) \cite{relu} for the hidden layers and sigmoid units for the output layer yielded the best results. Given the complexity of the label combinations in the training data, ReLU was best suited to be used for the model hidden layers given its simplicity and its ability to avoid the vanishing gradient problem encountered with sigmoid or tanh functions. The sigmoid function for output layer was chosen since the predicted output was binary for each label in the multi-label classification model. The model was then compiled with a binary cross entropy loss function, given that the prediction was a 0 or 1 for each class the verbatim was to be classified into. An optimal run of 50 epochs yielded the best combination of accuracy, F1 score, precision and recall based on empirical experimentation. Accuracy was computed using Keras's CategoricalAccuracy\footnote{\url{https://www.tensorflow.org/api_docs/python/tf/keras/metrics/CategoricalAccuracy}} metric whose value is calculated as the percentage of predicted labels that match the actual labels. Examples in Table \ref{tab:exampletable} indicate how each of the derived metrics were calculated. The total number of categories was 66, including 65 symptoms and 1 category for no symptom or 'unknown'. It is important to note that lower precision or lower recall doesn't necessarily indicate that the model has under performed. It is possible that the model actually detected patterns that were as yet not defined by the curation team and hence such verbatims should be flagged for further investigation. 

\begin{table}[h!]
\centering
\caption{Illustrative examples that demonstrate how we computed accuracy and other metrics for our multi-label classification experiments. (TP: True Positive, FP: False Positive, FN: False Negative, TN: True Negative)}. 
\begin{tabular}{|p{3.8cm}|p{4.2cm}|}
 \hline
 Exemplar True/Predicted Labels  & Metric Values\\
\hline
$Y_{true}$: ['balance']

$Y_{pred}$: ['falling'] & 

Categorical Accuracy: 0

TP: 0, TN: 65, FP: 0, FN: 1

Precision: 0, Recall: 0, F1: 0\\
\hline
$Y_{true}$: ['balance', 'falling']

$Y_{pred}$: ['balance'] &

Categorical Accuracy: 1

TP: 1, TN: 64, FP: 0, FN: 1

Precision: 1, Recall: 0.5, F1: 0.677\\
\hline
$Y_{true}$: ['balance'] 

$Y_{pred}$: ['balance', 'falling'] &

Categorical Accuracy: 0

TP: 1, TN: 64, FP: 1, FN: 0

Precision: 0.5, Recall: 1, F1: 0.677\\
\hline
$Y_{true}$: ['balance', 'falling']

$Y_{pred}$: ['balance', 'falling'] &

Categorical Accuracy: 1

TP: 2, TN: 64, FP: 0, FN: 0

Precision: 1, Recall: 1, F1: 1\\
\hline
\end{tabular}
\label{tab:exampletable}
\end{table}














\vspace{-3.5mm}
\section{Results}
The baseline model, i.e the model trained on the human annotated dataset of 1896 training samples, performed poorly on the held-out test set with low accuracy, recall and F1 scores, indicating the lack of sufficient training data. Precision on this model was still relatively high due to this being a multi-label classification and also due to over-fitting. Our proposed machine annotated model outperformed the baseline model on the held-out test on every metric as seen in Table \ref{tab:metricstable}. The accuracy improved by 42\% and recall improved by 79\% indicating the machine annotated model's efficiency in identifying specific multi-label combinations. To further eliminate the possibility of participant bias where the model overfits by learning from the same participant's symptom reporting patterns, we further examined a subset of the test set that only included those participants whose verbatims were not included in the train set (125 unique participants with sample size of 241 verbatims). 
We only observed a very small decrease in recall and F1 score when compared to the original machine annotated model, implying that the model could perform well on unseen data.

Overall, it is possible to improve precision further if the class imbalance on lower represented label combinations is improved as more data becomes available.

\begin{figure}[htb]
\centering
    \includegraphics[width=80mm]{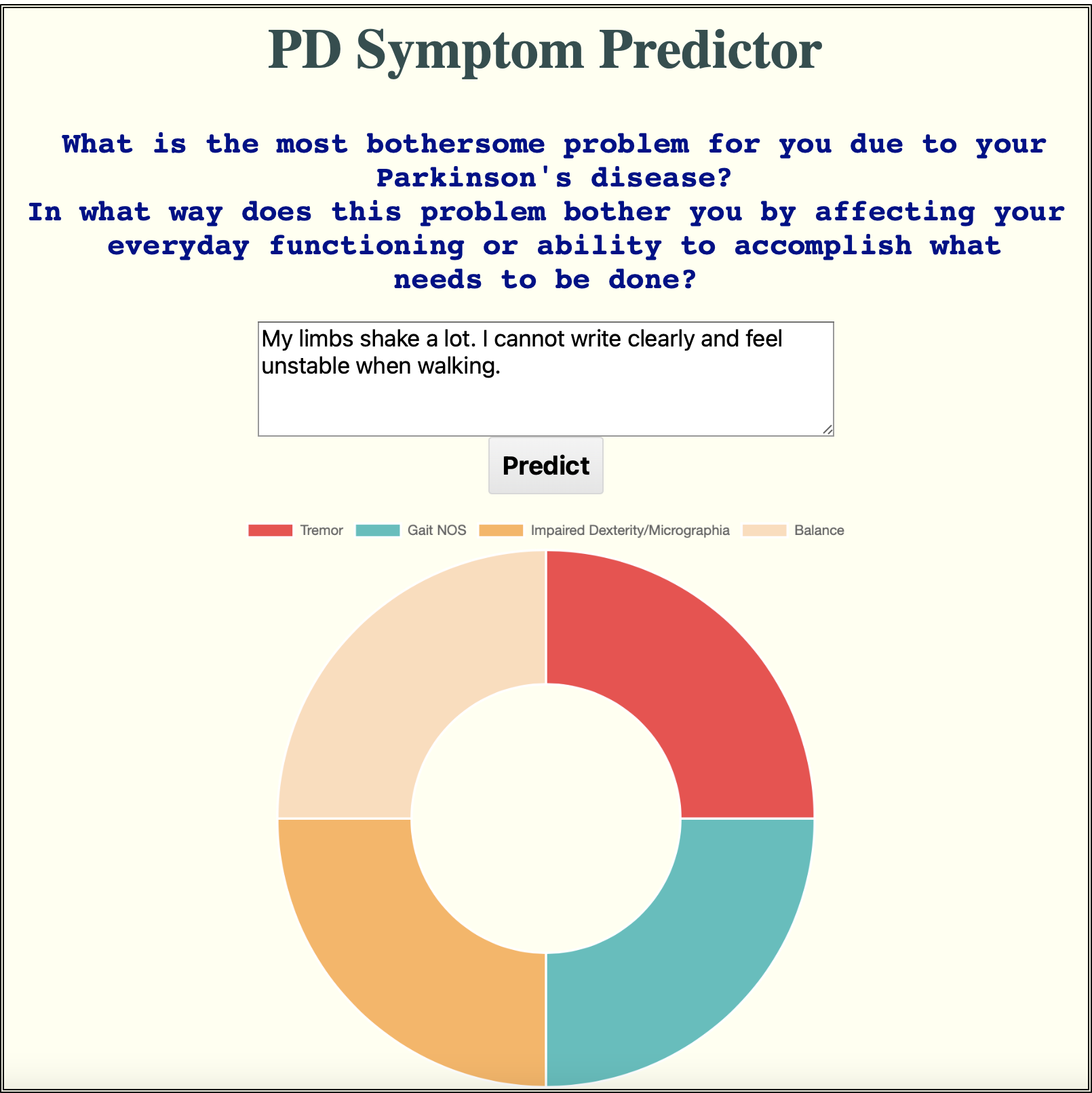}
    \caption{PD symptom predictor pilot interface}
    \label{fig:flask_image}
\end{figure}

\begin{table}[h!]
\centering
\caption{Metrics for baseline and Machine annotated ML models on the held-out test set}
\begin{tabular}{|p{1.3cm}|p{1.3cm}|p{1.5cm}|p{2cm}|}
 \hline
 Measure & Baseline model & Machine annotated model  & Machine annotated model (no participant bias)\\
\hline
Accuracy & 0.5217 & 0.9468 & 0.9499\\
\hline
F1-Score & 0.2407 & 0.9513 & 0.9499\\
\hline
Precision & 0.8125 & 0.9736 & 0.9877\\
\hline
Recall & 0.1413 & 0.9303 & 0.9160\\
\hline
\end{tabular}
\label{tab:metricstable}
\end{table}
\begin{figure}[htb]
    \includegraphics[width=0.5\textwidth,height=0.23\textheight]{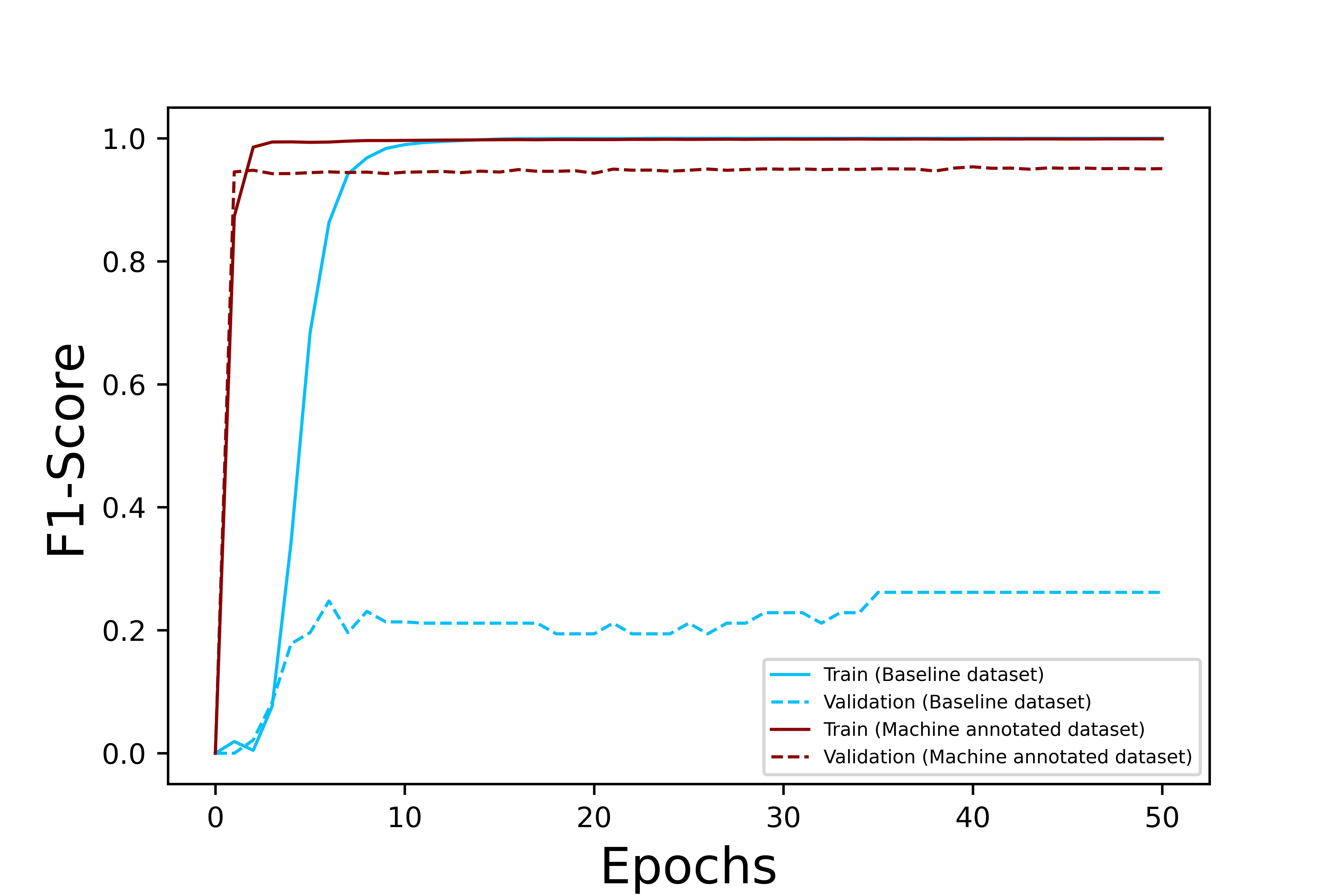}
    \caption{F1 scores for models trained on baseline and machine annotated datasets}
\end{figure}
\noindent This model was used in combination with the pre-processing layers to create an inference model that could directly predict PD symptoms on raw strings of input verbatim. To showcase the applications and capabilities of the inference model, a web application using a combination of the inference model, Python, Flask\footnote{\url{https://flask.palletsprojects.com}}, and Chart.js\footnote{\url{https://www.chartjs.org}} was used to create a web interface where real-time weighted PD symptom classifications could be displayed in graphical formats from user-input PD complaints. A snapshot of the pilot interface categorizing a verbatim is displayed in Figure \ref{fig:flask_image}. 

\section{Discussion and Next Steps}

This paper has presented a novel approach toward classifying patient verbatims into multiple, clinically relevant symptom categories. First, we built a baseline subset of verbatims annotated by nine human curators including clinical experts and PwPs. We then machine-annotated the remaining verbatims using a rules-based linguistic-dictionary and Neo4j based expert phrase-query system, and finally trained a Keras-Tensorflow model for  multilabel text classification. We further showed that the machine annotated model significantly outperformed the baseline model with a F1-score of 95\% across 65 symptom categories on a held-out test set. 

Our proposed modeling approach has multiple advantages. First, while the linguistic dictionary and the resulting model was built on the PD dataset, the defining feature of this model lies in its versatility in that it can be applied across other disease symptoms. This is because several motor and non-motor symptoms are disease agnostic. Second, building the symptom table by developing boundaries for each symptom via inclusion and exclusion criteria provides a clinical relevance to the derived output. The objective here was to reflect the voice of the patient by avoiding medical jargon, where possible, and to facilitate use of the symptom table by non-medical curation team members and future data users. Proposed boundaries were circulated among the curators and refined until consensus was reached. 

There are also multiple avenues of future work. 
While we showed that our model performed well even on verbatims from unseen participants, this was a small subset of our held-out test set. Going forward, we intend to evaluate the model's performance on a larger cohort of unseen participant data. 
Such an evaluation could also potentially inform new and as-yet undefined symptoms that can then be included in the symptom table, in an active machine learning framework, for further research and classification. 
Work is also underway to enrich the linguistic dictionary and build symptom tables for other diseases such as Huntington’s Disease and Multiple Sclerosis. With the extension of the linguistic dictionary and the symptom tables, the model could help evaluate pre-manifest and manifest disease conditions as well, further empowering researchers and clinicians to administer appropriate and timely care to patients. 


\section*{Acknowledgements}

This work was funded in part by a grant from the Michael J Fox Foundation. The authors also thank the FI participants for providing their responses to the PD-PROP questionnaire.

\bibliographystyle{IEEEtran}
\bibliography{what_do_patients_say_arXiv2023}




\begin{table*}[h]
\centering
\caption{Validation metrics for symptoms extracted by Neo4j process}
\begin{tabular}{ |p{4cm}|p{7cm}|p{2cm}| }
 \hline
 \textbf{Domain} & \textbf{Symptom} & \textbf{F1-Score}\\
 \hline Tremor & tremor & 1 \\\cline{2-3}
  & internal tremor & 1 \\
\hline Ridigity & stiffness & 1 \\
\hline Bradykinesia & slowness & 0.9714 \\\cline{2-3}
  & facial expression & 1 \\
\hline Postural Instability & Balance & 1 \\\cline{2-3}
  & Falls & 1 \\\cline{2-3}
  & Fear of falling & 1 \\
\hline Gait & Gait not otherwise specific & 0.9859 \\\cline{2-3}
  & Freezing of gait & 1 \\
\hline Other Motor & Impaired Dexterity & 1 \\\cline{2-3}
  & Dystonia & 1 \\\cline{2-3}
  & Posture & 1 \\\cline{2-3}
  & speech & 1 \\
\hline Fluctuations & Off Periods—medication related & 1 \\\cline{2-3}
  & Off Periods—medication not mentioned & 1 \\\cline{2-3}
  & Random off/unpredictable off & 1 \\\cline{2-3}
  & Medications not working nos & 1 \\
\hline Dyskinesias & Dyskinesias & 1 \\
\hline Sleep & Excessive Daytime Sleepiness (ES) & 1 \\\cline{2-3}
  & Sleep onset insomnia & 1 \\\cline{2-3}
  & Sleep Maintenance Insomnia & 1 \\\cline{2-3}
  & Early Morning Awakening & 1 \\\cline{2-3}
  & Poor sleep quality NOS & 1 \\\cline{2-3}
  & RLS/restlessness & 1 \\\cline{2-3}
  & RBD -like symptoms & 1 \\\cline{2-3}
  & Parasomnia Unspecified & 0.9859 \\\cline{2-3}
  & Dreams & 1 \\
\hline Fatigue & Physical Fatigue & 1 \\\cline{2-3}
  & Mental fatigue & 1 \\
\hline Cognition & Memory & 0.9565 \\\cline{2-3}
  & Concentration/Attention & 1 \\\cline{2-3}
  & Cognitive Slowing/Mental fatigue & 0.9565 \\\cline{2-3}
  & Language/Word Finding & 1 \\\cline{2-3}
  & Mental alertness/awareness & 1 \\\cline{2-3}
  & Visuospatial abilities & 0.9565 \\\cline{2-3}
  & Executive abilities/working memory & 1 \\\cline{2-3}
  & Cognitive impairment not otherwise specified & 1 \\
\hline Affect/Motivation/Thought-Perception/Other Psychiatric & Depressive symptoms & 0.9524 \\\cline{2-3}
  & Death and Suicidal ideation & 1 \\\cline{2-3}
  & Apathy & 1 \\\cline{2-3}
  & Anxiety/Worry & 1 \\\cline{2-3}
  & Delusions/Psychosis & 1 \\\cline{2-3}
  & Loneliness/isolation & 1 \\\cline{2-3}
  & Impulse control & 1 \\\cline{2-3}
  & Hallucinations/Illusion/Presence/Passage & 1 \\\cline{2-3}
  & Pseudobulbar affect & 1 \\\cline{2-3}
  & Negative emotions not otherwise specified & 1 \\
\hline Pain & Pain/Discomfort & 1 \\\cline{2-3}
  & Cramp or spasm & 1 \\\cline{2-3}
  & Headache & 1 \\
\hline Autonomic Dysfunction & Bowel incontinence & 1 \\\cline{2-3}
  & Bowel urgency & 1 \\\cline{2-3}
  & Bloating/Feeling Full & 1 \\\cline{2-3}
  & Altered Bowel frequency & 1 \\\cline{2-3}
  & Abdominal  discomfort not otherwise specified & 1 \\\cline{2-3}
  & Bladder incontinence & 1 \\\cline{2-3}
  & Excessive Sweating & 1 \\\cline{2-3}
  & Frequent Urination & 1 \\\cline{2-3}
  & Lightheadedness/dizziness with a change in posture & 1 \\\cline{2-3}
  & Swallowing problems & 1 \\\cline{2-3}
  & Temperature dysregulation & 1 \\\cline{2-3}
  & Sexual dysfunction & 1 \\\cline{2-3}
  & Diarrhea & 1 \\\cline{2-3}
  & Nausea & 1 \\
 \hline
\end{tabular}
\label{tab:validation_metrics}
\end{table*}

\end{document}